\documentclass[letterpaper, 10 pt, conference]{ieeeconf}  

\IEEEoverridecommandlockouts                              

\usepackage{color}
\usepackage{epsfig}
\usepackage{graphicx}
\usepackage{algorithm,algorithmic}

\usepackage{adjustbox}
\usepackage{array}
\usepackage{booktabs}
\usepackage{colortbl}
\usepackage{float,wrapfig}
\usepackage{framed}
\usepackage{hhline}
\usepackage{multirow}
\usepackage{makecell}
\usepackage[percent]{overpic}

\usepackage{amsmath,amsfonts,amssymb}
\usepackage{amsthm} 
\usepackage{bm}
\usepackage{nicefrac}
\usepackage{microtype}
\usepackage{contour}
\usepackage{courier}
\usepackage{siunitx}

\usepackage{changepage}
\usepackage{extramarks}
\usepackage{fancyhdr}
\usepackage{lastpage}
\usepackage{setspace}
\usepackage{soul}
\usepackage{xspace}
\usepackage{cuted}
\usepackage{fancybox}
\usepackage{afterpage}
\usepackage{gensymb}

\usepackage[breaklinks=true,colorlinks,backref=True]{hyperref}
\hypersetup{colorlinks,linkcolor={black},citecolor={MSBlue},urlcolor={magenta}}
\usepackage{url}
\usepackage{quoting}
\usepackage{epigraph}

\usepackage{enumerate}
\usepackage{paralist,tabularx}
\usepackage{comment}
\usepackage{pdfpages}
\usepackage{caption}  

\usepackage{pifont}

\usepackage{MnSymbol}

\usepackage{bbold}

\usepackage{footmisc}   
\usepackage{cite}

\makeatletter
\DeclareRobustCommand\onedot{\futurelet\@let@token\@onedot}
\def\@onedot{\ifx\@let@token.\else.\null\fi\xspace}

\makeatother

\definecolor{MyDarkBlue}{rgb}{0,0.08,1}
\definecolor{MyDarkGreen}{rgb}{0.02,0.6,0.02}
\definecolor{MyDarkRed}{rgb}{0.8,0.02,0.02}
\definecolor{MyDarkOrange}{rgb}{0.40,0.2,0.02}
\definecolor{MyPurple}{RGB}{111,0,255}
\definecolor{MyRed}{rgb}{1.0,0.0,0.0}
\definecolor{MyGold}{rgb}{0.75,0.6,0.12}
\definecolor{MyDarkgray}{rgb}{0.66, 0.66, 0.66}
\definecolor{MyPink}{rgb}{1, 0.75, 0.79}
\definecolor{GreenStarColor}{rgb}{0.54, 0.84, 0.41}
\definecolor{MSBlue}{rgb}{0, 0.35, 0.49}
\definecolor{Cardinal}{rgb}{0.72, 0.23, 0.29}

\definecolor{policycolor}{rgb}{0.576,0.769,0.490}
\definecolor{interventioncolor}{rgb}{0.427,0.620,0.922}

\def\OURS{DITTO-X\xspace}

\usepackage{xcolor}
\definecolor{ditogreen}{RGB}{0,128,0}
\definecolor{ditored}{RGB}{190,30,45}
\newcommand{\yes}{\textcolor{ditogreen}{$\checkmark$}}
\newcommand{\no}{\textcolor{ditored}{$\times$}}

\usepackage[normalem]{ulem}

\title{\huge \bf Forward and Reverse Teleoperation for \\ Dexterous Manipulation and Human Intervention}

\author{
Zhanpeng He\textsuperscript{*1}, Joaquin Palacios\textsuperscript{*2}, Zhangyu Wang\textsuperscript{1}, Chenhao Li\textsuperscript{1}, Katelyn Lee\textsuperscript{2} \\%
Matei Ciocarlie\textsuperscript{\dag2}, C. Karen Liu\textsuperscript{\dag1}, Jiajun Wu\textsuperscript{\dag1}%
}

\begin{document}

\bstctlcite{IEEEexample:BSTcontrol}

\twocolumn[{%
\renewcommand\twocolumn[1][]{#1}%
\maketitle
\vspace{-1.4em}
{\centering
{\small \href{https://tml.stanford.edu/ditto-x/}{\textbf{\texttt{\textcolor{Cardinal}{https://tml.stanford.edu/ditto-x/}}}}}\par}
\vspace{0.4em}

\begin{center}
    \centering
    \captionsetup{type=figure, font=small}
    \includegraphics[width=1.0\textwidth]{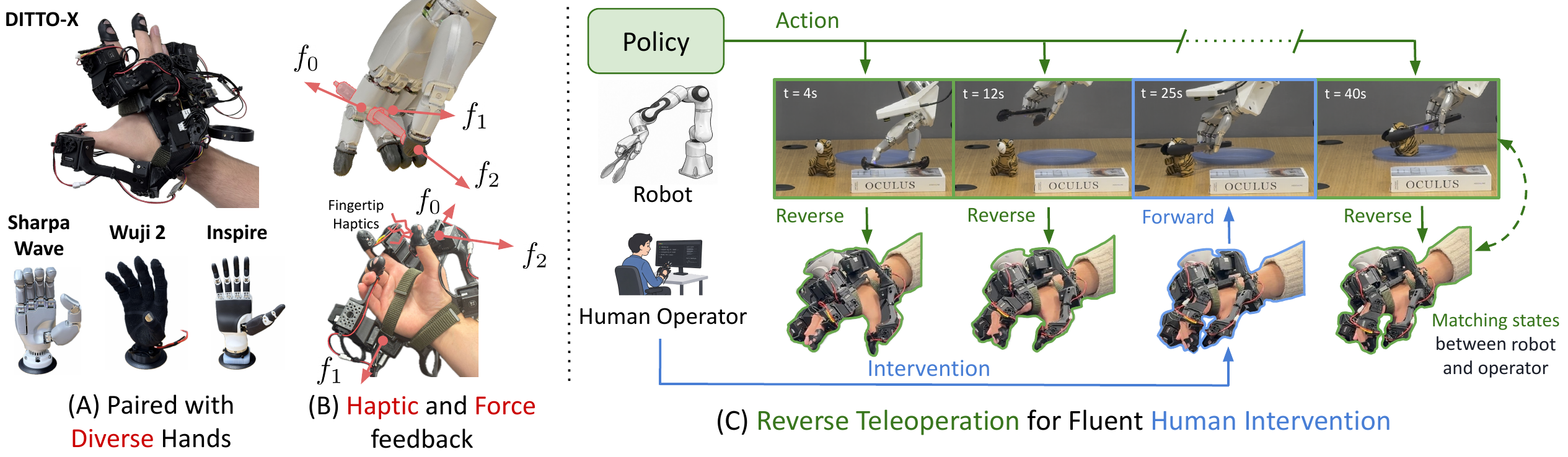}
    \caption{\textbf{Overview of \OURS.} \OURS supports \textit{forward teleoperation} of diverse dexterous hands (A) with joint-level force and fingertip haptic feedback (B). During policy deployment, \OURS also supports \textit{reverse teleoperation} (C), in which the robot hand drives the human hand to match its state, enabling fluent human intervention. As a result, a human operator can interleave the two directions (C) to complete contact-rich manipulation tasks under shared autonomy. }
    \label{fig:candy}
\end{center}
}]
\newcommand{\authornotefootnote}[1]{%
  \begingroup
  \renewcommand{\thefootnote}{}%
  \footnotetext{#1}%
  \endgroup
  \addtocounter{footnote}{-1}%
}
\authornotefootnote{$^{*}$ Equal contribution. \quad $^{\dagger}$ Equal advising. $^{1}$ Department of Computer Science, Stanford University, Stanford, CA 94305, USA. $^{2}$ Department of Mechanical Engineering, Columbia University, New York, NY 10027, USA.}

\thispagestyle{empty}
\pagestyle{empty}


\begin{abstract}
Teleoperated demonstrations are a primary source of data for robot manipulation, and teleoperated interventions are a primary mechanism for correcting policies at deployment. Yet most teleoperation systems close the loop through vision alone and are built around parallel-jaw grippers, limiting both what the robot can execute and what the operator can express through it. 
This is most damaging in shared autonomy, where the operator sees the scene only through occluded cameras and must take over a dexterous hand mid-task, often with an object already grasped. 
We present \OURS, a hand-agnostic dexterous teleoperation interface that renders joint-level force and fingertip contact events from sensing already on the robot hand, and drives three commercial dexterous hands (Sharpa, Wuji, and Inspire) without per-hand redesign. Because the exoskeleton is actuated, \OURS also supports \textit{reverse teleoperation}, in which the robot back-drives the operator's fingers into its own configuration before control is transferred, so the human enters the loop already matched to the state they inherit. Our results show that \OURS improves demonstration quality and throughput over a commercial hand-tracking glove, both in regular data collection and in human intervention during policy deployment for contact-rich manipulation tasks.

\end{abstract}

\section{Introduction}
\label{sec:intro}
Teleoperation plays a fundamental role in modern robotics, letting humans directly supervise, guide, and demonstrate complex manipulation tasks~\cite{intelligence2025pi, bjorck2025gr00t}. In robot learning, teleoperated demonstrations remain a critical complement to scripted motion, human video, and handheld collection devices~\cite{chi2024universal, fang2025dexop}, capturing the precise, contact-rich interactions that passive or ungrounded approaches struggle to reproduce. At deployment time, shared autonomy and human-in-the-loop frameworks rely on teleoperation to let operators intervene at uncertain or safety-critical moments~\cite{he2025uncertainty, hoque2021thriftydagger, agia2024unpacking}, and the interventions they produce can be used to improve the policy.

Despite their widespread use, most teleoperation systems are poorly matched to dexterous manipulation. They are built around parallel-jaw grippers, which bound the natural movement an operator can demonstrate in the first place, and they close the loop through vision alone. Humans do not manipulate this way; we regulate contact forces, detect incipient slip, and reshape a grasp from cues we feel rather than see. Without those cues, the operator is blind to contact, and demonstrations are rendered slow and conservative. The limitation is therefore not only what the robot can execute, but also what the human can express through it.

This limitation becomes a failure mode in \textit{shared autonomy}. When collecting demonstrations from scratch, the operator can both position themselves to see the scene and restart a failed attempt at little cost. Intervening on a running policy affords neither. Perception is restricted to onboard cameras, under which self-occlusion by the arm and hand is often unavoidable, especially at critical contact moments. The operator also inherits the robot's state mid-task, so the kinematic mismatch between the human hand and the robot hand must be resolved at that exact moment. During tool use or in-hand manipulation, a non-smooth transition at takeover is enough to drop or break the object, resulting in unrecoverable failures.

We present \OURS, a dexterous interface for transparent teleoperation that works across different robotic hands for both regular data collection and human intervention. Improving the design of DITTO~\cite{ditto}, which requires strict 1-to-1 kinematics between leader and follower and instruments only two fingers, \OURS is a hand-agnostic framework and currently drives three commercial, five-finger dexterous hands (Sharpa Wave, Wuji 2, and Inspire RH56DFX) without per-hand redesign. It closes the feedback loop at two levels: sustained joint-level force via motors on the exoskeleton and discrete fingertip contact events via haptics from sensing already available on the robot hand. This allows the operator to feel both when contact is made and how hard the robotic hand is squeezing, restoring cues that vision-only interfaces lack.

Our actuated exoskeleton can also be used to improve human intervention through a novel operating mode we call \textit{``reverse teleoperation''}: while an operator is supervising an autonomous policy deployment, the exoskeleton can actively \textit{drive} the operator's fingers to match the robot's current configuration. This allows for a seamless transition of control once the operator takes over and gives them a kinesthetic sense of the policy actions.

Our user study and policy learning results show that \OURS improves both the throughput and the quality of data collected for contact-rich policy training. With reverse teleoperation, operators intervene on a running policy fluently, without dropping objects already in hand, yielding better data for subsequent policy improvement.
We summarize our contributions as follows:
\begin{itemize}
    \item We present a dexterous teleoperation framework, equipped with force and haptic feedback, with support for three commercial robotic hands, spanning 6 to 22 actuated DoF, without hand-specific redesign.
    \item We introduce \textit{reverse teleoperation}, in which the actuated exoskeleton back-drives the operator's hand to match the robot's configuration, so that intervention on a running policy begins from a matched state rather than a discontinuous handoff.
    \item We open-source the full system, including hardware design, forward and reverse teleoperation, and policy training and fine-tuning details. We also release all 1.8k episodes of contact-rich manipulation data from three embodiments, spanning teleoperated demonstrations, human interventions, and autonomous policy rollouts.
\end{itemize}
\section{Related Work}
\label{sec:related}

\textbf{Teleoperation for dexterous hands.} Dexterous teleoperation has largely been studied as a problem of closing the kinematic gap between the operator's hand and the robot's: vision-based hand tracking~\cite{qin2023anyteleop}, non-wearable leader-follower designs~\cite{tilde}, and learned retargeting priors~\cite{dexterity_gen} each attack it differently, but none renders contact back to the operator, who is left to infer it from vision alone.
Motorized exoskeletons do render force, yet packaging actuators alongside the fingers forces a choice about what to render. DOGlove~\cite{doglove} reduces feedback to a single axis per finger, while GEX~\cite{gex} and N2D~\cite{huang2026n2d} map forces in task space; neither tells the operator which joint is loaded. DITTO~\cite{ditto} recovers joint-level feedback by enforcing a one-to-one correspondence between leader and follower joints, at the price of being tied to a single two-fingered manipulator. In this work, we remove that restriction and extend the design to five-fingered hands.

\textbf{Shared autonomy, human intervention and DAgger.} A large body of work improves manipulation policies with human feedback, including interventions~\cite{mandlekar2020human, spencer2020learning}, preferences~\cite{lee2021pebble}, rankings~\cite{pmlr-v97-brown19a}, and scalar rewards~\cite{macglashan2017interactive}. Interactive imitation learning, in which the agent queries an expert during execution and aggregates the resulting corrections~\cite{ross2011reduction, kelly2019hg}, underpins systems that collect data where the policy actually fails~\cite{liu2022robot, luo2024hilserl}. Because continuous supervision is costly, much of this literature concerns when to request help~\cite{hoque2023fleet, hoque2021thriftydagger, agia2024unpacking, lee2024diff, zhao2025conformalized}, and a parallel line supplies it through language rather than teleoperation~\cite{cui2023lilac, shi2024yay}. Most of these works study when a human should intervene. In this work, we look into the problem of how to intervene on a policy.

How authority passes between an autonomous controller and a human has also been studied. Blended and traded control arbitrate continuously, predicting the operator's goal and blending toward it~\cite{dragan2013policy, javdani2015shared} or applying guidance forces that shift authority without a discrete switch~\cite{abbink2012haptic}; when a switch is unavoidable, the resulting transient is well documented in automated driving~\cite{eriksson2017takeover}. Our setting differs in what must be made continuous. A dexterous hand transferred mid-task is usually already holding an object, so a pose mismatch at the instant of transfer releases or displaces it rather than merely perturbing a trajectory. Reverse teleoperation removes this discontinuity in the operator's state rather than the controller's, conforming the human to the robot before authority passes.

\section{Method}
\label{sec:method}

\begin{figure}[t]
    \centering
    \includegraphics[width=\columnwidth]{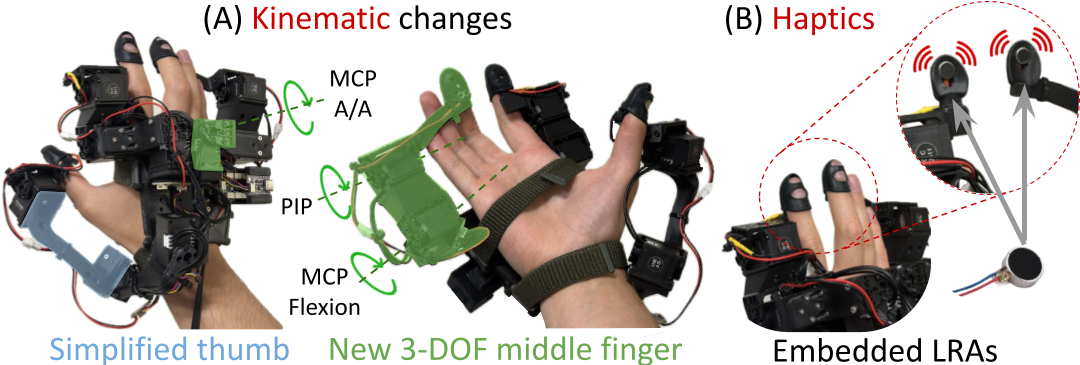}
    \caption{Design improvements from DITTO. \textbf{(A)} A third finger module instruments middle-finger MCP abduction/adduction, MCP flexion, and PIP flexion, and a simplified thumb reduces weight. \textbf{(B)} Fingertip haptics from 8\,mm z-axis LRAs (JYLRA0825Z) driven by DRV2605L drivers. \vspace{-2em}}

    \label{fig:ditto_hardware_updates}
\end{figure}

\OURS supports two modes of operation: \textit{forward} teleoperation (human-to-robot) and a novel \textit{reverse} teleoperation (robot-to-human). In the \textit{forward} direction, used for data collection, it transmits the operator's hand motion to a dexterous robot hand while rendering the contact the robot experiences back to the operator via force and haptic feedback. In the \textit{reverse} direction, which is intended for intervention during policy deployment, it drives the operator's fingers to match the robot's configuration, which enables a fluent transition when the operator takes over. 

We first describe how \OURS supports high-quality \textit{forward} teleoperation across different robot hands, augmented by force and haptic feedback (Sections~\ref{sec:forward} and~\ref{sec:feedback}). We then describe how we leverage the same actuation to support fluent takeover via \textit{reverse} teleoperation (Section~\ref{sec:reverse}).

\subsection{Forward Teleoperation Across Dexterous Hands}
\label{sec:forward}

\OURS is designed to support bilateral teleoperation for a diverse set of robotic hands, with vastly different kinematic designs and sensing capabilities. We select DITTO as a starting point due to its kinematic design of thumb and index fingers, which aligns with the user's anatomical joint axes, allowing for joint-level position control and force feedback. However, DITTO relies on a kinematically equivalent leader-follower pair, which limits its applications to other manipulators and five-finger hands. 

To address these problems, we extend its design by adding a third motorized finger module interfacing with the user's middle finger, which mirrors the index design to the ulnar side of the hand, aligning actuators with the middle finger's joint axes (Fig. \ref{fig:ditto_hardware_updates}.A Left). To support five-finger hand control, we couple the motion of the ring and pinky fingers on the target hand with the middle finger, positing that three independent fingers suffice for a variety of dexterous primitives, while the coupled ring and pinky fingers allow for power grasps (Fig.~\ref{fig:ditto_hardware_updates}). We also remove one joint from the original thumb design (3rd joint from the base, i.e. proximal flexion joint) to reduce weight (Fig. \ref{fig:ditto_hardware_updates}.A right); we find this removal does not degrade our ability to capture thumb motion. Additionally, we expand the modalities of contact feedback by embedding Linear Resonant Actuators (LRAs) (Fig.~\ref{fig:ditto_hardware_updates}B) underneath the operator's fingertips for vibro-tactile feedback.

We use \OURS on three different robotic hands: Sharpa, Wuji2, and Inspire with vastly different kinematic designs and sensing capabilities (Table \ref{tab:hand_comparison_expanded}). To support this heterogeneity, we select both the position mapping and the force and haptic feedback for each hand according to its kinematics and available sensing.

\subsubsection{Position Retargeting under Heterogeneous Kinematics}
\label{sec:position_control}
Bilateral teleoperation between the operator's hand and the target hand requires bridging their kinematic gap, for both position control and force feedback.
Because \OURS aligns its motor axes with the operator's anatomical joint axes, the index and middle fingers admit a one-to-one joint mapping: MCP abduction, MCP flexion, and PIP flexion on \OURS pair directly with the corresponding joints on the target hand, a common arrangement across anthropomorphic hands. For position control, we send each \OURS joint angle to its counterpart and hold unmatched joints at a neutral angle.

For the thumb, however, more kinematic diversity exists among off-the-shelf hands, which requires retargeting in task space. We define palm reference frames fixed at the index MCP joint of both \OURS and the target hand. Forward kinematics gives the \OURS thumb-tip pose in its own palm frame, which we express in the target-hand palm frame through a fixed transform between the two (scaled by a per-hand factor $s_f$ that accounts for differences in thumb length), and solve inverse kinematics for the target-hand thumb joints. The same palm-frame definitions carry over to force retargeting in Section~\ref{sec:feedback}. For underactuated fingers, such as those on Inspire, we instead map total \OURS finger flexion to a single target-hand flexion DoF.

\subsubsection{Force and Haptic Feedback under Heterogeneous Sensing}
\label{sec:feedback}

To render force feedback, we must first sense the forces experienced by the target hand. The supported hands provide either motor-current measurements or dedicated fingertip tactile sensing (Table \ref{tab:hand_comparison_expanded}). Motor currents yield estimates in the target hand's joint space, whereas tactile sensors directly report a fingertip force vector. For each finger, we use these measurements to estimate a set of desired feedback torques $\hat{\tau}^{\mathrm{des}}_l$, one torque $\hat{\tau}^{\mathrm{des}}_{l,j}$ per DITTO-X actuator $j$.

\textbf{Motor-current sensing with 1:1 joint mapping.}
We convert the measured target-hand motor currents into joint-torque estimates $\hat{\tau}_f$. When the target-hand joints have a 1:1 mapping to DITTO-X joints, as for the index and middle fingers on Sharpa and Wuji, each estimated target-hand joint torque directly becomes the desired feedback torque of the corresponding DITTO-X joint, similar to joint-level force feedback in~\cite{ditto}.

\begin{table}[t]
\centering
\caption{Comparison of supported robotic hands. DITTO-X 1:1 Joint Mapping indicates whether a finger on the target hand can be paired joint-to-joint directly with the corresponding finger on DITTO-X.}
\label{tab:hand_comparison_expanded}
\small
\setlength{\tabcolsep}{3pt}
\begin{tabular}{llccc}
\toprule
& & \textbf{Sharpa} & \textbf{Wuji2} & \textbf{Inspire} \\
\midrule
\multicolumn{2}{l}{\textbf{Actuated DoF}} & 22 & 20 & 6 \\
\cmidrule(lr){1-5}
\multirow{2}{*}{\textbf{\makecell[l]{DITTO-X 1:1\\Joint Mapping}}}
    & Thumb         & \no  & \no  & \no\ (2 DOF only) \\
    & Index, Middle & \yes & \yes & \no\ (1 DOF only) \\
\cmidrule(lr){1-5}
\multirow{2}{*}{\textbf{Sensing}}
    & Joint Current & \yes & \yes & 1D / finger \\
    & Tactile       & \yes & \no  & \no \\
\cmidrule(lr){1-5}
\multirow{2}{*}{\textbf{Feedback}}
    & Torque        & \yes & \yes & 1D / finger \\
    & Vibration     & \yes & \no  & \no \\
\bottomrule
\vspace{-3.5em}
\end{tabular}
\end{table}
\begin{figure*}[t]
    \centering
    \setlength{\abovecaptionskip}{2pt}
    \includegraphics[width=\linewidth]{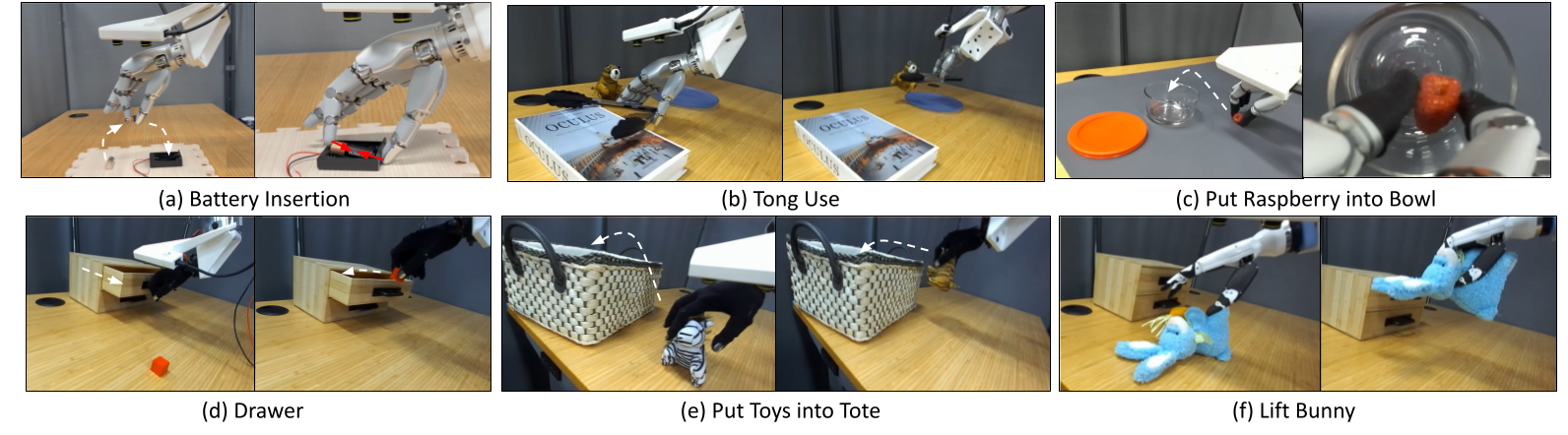}
    \caption{\textbf{Evaluation tasks.} \textbf{(a)}~Battery, \textbf{(b)}~tong, and \textbf{(c)}~raspberry on the Sharpa hand span forceful insertion, tool use, and fragile grasping. \textbf{(d)}~Drawer, on Wuji: one episode recruiting different parts of the hand in turn, hooking, pinching, then closing with a fist. \textbf{(e)}~Bunny lifting, on Inspire, a hand with only six actuated degrees of freedom.\vspace{-2.5em}}
    \label{fig:tasks}
\end{figure*}

\textbf{Motor-current sensing without 1:1 joint mapping.}
Where joint correspondence is unavailable, as for the thumb, we retarget force through task space. Since $\tau_f = J_f^\top F_f$, we estimate the target-hand fingertip force as $\hat{F}_f = (J_f^\top)^\dagger \hat{\tau}_f$, with $J_f$ the target-hand fingertip position Jacobian and $\dagger$ the pseudoinverse. Expressing this in the \OURS fingertip frame using the palm frames of Section~\ref{sec:position_control} and applying the Jacobian transpose gives:
\begin{equation}
\hat{\tau}^{\mathrm{des}}_l = J_l^\top R_l R_f^\top \hat{F}_f,
\end{equation}
where $J_l$ is the \OURS fingertip position Jacobian and $R_f$, $R_l$ are the target-hand and \OURS fingertip-frame orientations. This produces the desired feedback torques in \OURS joint space. Inspire reports one force magnitude per finger rather than joint current, which we treat as a fingertip force and map through the same Jacobian transpose.

\textbf{Tactile sensing.}
When the target hand provides a fingertip force vector directly, as on Sharpa, we can skip the joint-to-task-space conversion. We instead express the tactile force in the DITTO-X fingertip coordinate frame and apply the same DITTO-X Jacobian-transpose mapping above to obtain $\hat{\tau}^{\mathrm{des}}_l$.

\textbf{Rendering the desired feedback torques.}
Once we obtain the desired feedback torques $\hat{\tau}^{\mathrm{des}}_l$, we low-pass filter each one to obtain $\tilde{\tau}^{\mathrm{des}}_{l,j}$ and render a proportional motor current per actuator using the DITTO control law:
\begin{equation}
\label{eq:force_law}
i^{\mathrm{cmd}}_{l,j} =
\begin{cases}
-k_j\,\mathrm{sign}\!\left(\tilde{\tau}^{\mathrm{des}}_{l,j}\right)
\left(\left|\tilde{\tau}^{\mathrm{des}}_{l,j}\right|-\tau_{\mathrm{th},j}\right),
& \left|\tilde{\tau}^{\mathrm{des}}_{l,j}\right| > \tau_{\mathrm{th},j}, \\
0, & \text{otherwise,}
\end{cases}
\end{equation}
where the tunable gain $k_j$ bridges the scale between the retargeted torque estimate and the commanded actuator current. For torques estimated from motor current, $\tau_{\mathrm{th},j}$ rejects estimation bias during free-space motion. For torques obtained from tactile sensing, we set $\tau_{\mathrm{th},j}=0$ because the tactile sensor directly measures external contact force.

\textbf{Haptic contact events.}
While force feedback can effectively convey sustained contact forces, we find that it requires high filtering for stability, making it less suitable for fast-changing contact events. We therefore leverage LRAs at the user's fingertips to convey high-frequency contact events as vibrotactile feedback (Fig.~\ref{fig:ditto_hardware_updates}B). 

This modality requires contact event estimation, which we support only for hands with fingertip tactile sensing (i.e.\ Sharpa). We estimate contact state $c_t$ from the tactile force magnitude $|F|$. Because these sensors exhibit hysteresis under loading versus unloading, we use asymmetric thresholds selected by the rate of change of $|F|$ (rising versus falling):
\begin{equation}
c_t =
\begin{cases}
1, & |F| \ge F_t, \\
0, & |F| < F_t,
\end{cases}
\qquad
F_t =
\begin{cases}
F_{\mathrm{rising}}, & \mathrm{d}|F|/\mathrm{d}t > 0, \\
F_{\mathrm{falling}}, & \mathrm{d}|F|/\mathrm{d}t < 0.
\end{cases}
\end{equation}
where $F_{\mathrm{rising}} < F_{\mathrm{falling}}$ such that the lower rising threshold detects light onsets, while the higher falling threshold requires a clear unload before contact is cleared.
We find it best to render only contact onset, as prolonged vibration only confuses the operator: on the rising edge of $c_t$, we map $|F|$ to a vibration intensity and then decay to zero after a time $t_d$ (we find $0.5$\,s to be suitable). After decay, the vibrotactile feedback stays off until $c_t$ returns to zero, after which it can be retriggered by a new contact event. Vibration therefore marks the contact event, while sustained grasp force continues to be rendered by joint-torque feedback.

\subsection{Reverse Teleoperation for Fluent Human Intervention}
\label{sec:reverse}

The previous subsection describes \OURS in the context of forward teleoperation, where the operator holds \textit{full authority} over the robot, as when gathering training data. During robot deployment, teleoperation can be used for human intervention of an autonomous system (e.g., a learning policy) to correct any undesired behaviors. These interventions can further be aggregated for policy improvement~\cite{ross2011reduction}, leading to more robust autonomous performance. 

However, with dexterous hands, takeover is challenging: the large joint space makes it hard for the operator to match the robot's current configuration, introducing a discontinuity during handoff that can drop or crush a grasped object and fail the task. \OURS addresses this via \emph{reverse teleoperation}: while the policy is in control, we project the robot hand's commanded configuration back onto the exoskeleton, so the operator's hand stays matched to the robot and the transition at takeover is seamless.

\textbf{Projecting the policy command.} The projection inverts whichever forward mapping each finger uses. When a 1:1 joint correspondence exists, the inverse is exact and requires no solver. When retargeting is needed, as for the thumb, we run it backwards: forward kinematics on the target hand gives its fingertip pose, which we express in the \OURS palm frame, rescale by $1/s_f$, and solve for the \OURS joints that place the fingertip there.

\textbf{Takeover.} The operator claims control with a keyboard command, but control is held until the measured \OURS configuration departs from the projection by more than a threshold $\delta_{\mathrm{release}}$ on any joint, so an accidental keypress changes nothing and the transfer begins only once the operator has actively moved. 
This gate depends on the projection remaining an independent reference: we therefore seed the inverse-kinematics solve from the previous projection rather than from the measured \OURS configuration, since a redundant solve re-seeded from a displaced hand would follow that hand, shrink the measured deviation, and conceal the very motion the threshold is meant to detect. Only the first solve is seeded from the measured pose.

\textbf{Handover.} Control returns to the policy through the same trigger, so an intervention need not run to the end of the episode. An operator can therefore interleave forward and reverse teleoperation to finish a task.

\textbf{Safety Consideration.} During both force-feedback and reverse-teleoperation modes, we ensure user safety via: software joint limits within a comfortable range of motion, preventing hyperextension, conservative motor current limits ranging under $200$\,mA depending on the joint ($\leq$0.08\,N$\cdot$m), and an emergency e-stop.

\section{Experiments}

\label{sec:exp}
To evaluate our method, we design our user study and policy learning experiments to answer the following questions.
\textbf{Q1}: Can \OURS provide useful force information and improve throughput and success rates for collecting dexterous, forceful task data?
\textbf{Q2}: Does reverse teleoperation improve human intervention of a learned policy and obtain high-quality data for continuous policy improvement?
\textbf{Q3}: Can \OURS be used for diverse, dexterous hands?

\subsection{User Study}
To evaluate both the forward teleoperation and human intervention capabilities of \OURS, we conducted a user study ($n=6$) spanning the four tasks listed below.

\begin{figure}[t]
    \centering
    \setlength{\abovecaptionskip}{0pt}
    \includegraphics[width=\linewidth]{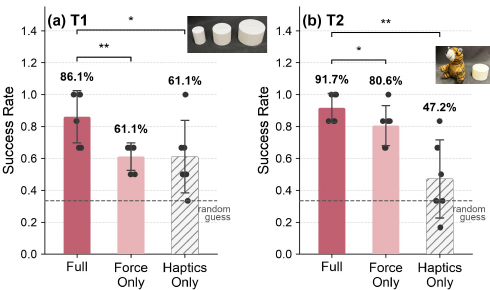}
    \caption{Object size (T1) and compliance (T2) identification accuracy. Objects used for each task are shown on the right of the corresponding subfigure.\vspace{-3em}}
    \label{fig:user_e1_e2}
\end{figure}

\textbf{T1: Object size discrimination without vision.} We place a cylinder of one of three diameters (30, 50, or 70\,mm), shown in Fig.~\ref{fig:user_e1_e2}, in the Sharpa hand and ask the subject to close a grasp with their eyes closed. The subject reports which size they are holding from the rendered feedback alone. Each subject performs 6 trials per condition.

\textbf{T2: Object compliance discrimination without vision.} As in T1, but the hand holds a soft object, a rigid object, or nothing at all. Including the empty case tests whether the interface conveys the presence of contact and not only its character. Again 6 trials per condition, with the same metric.

\textbf{T3: Tool-use data collection.} The subject completes 10 trials per condition of a tool-use task: pick up a pair of tongs, grasp a toy with them, and place it in a target region without dropping the tongs. We compare \OURS against Manus Pro (Manus hereafter), a commercial glove that tracks hand pose through magnetically sensed finger positions. It provides no feedback and cannot drive the operator's hand, so it serves as a pure-tracking baseline that isolates the contribution of the bilateral loop. We report success rate and time per success.

\textbf{T4: Intervening on a learned tool-use policy.} We deploy a tool-use policy trained on \OURS demonstrations using the setup of Section~\ref{sec:policy_data}, and ask the subject to take over once the robot has picked up the tongs, completing the remainder of the task as in T3. Each subject performs 10 trials per condition. We measure success rate on the full task, counting a dropped tool as a failure: recovering from a dropped tool requires resetting the scene, which is precisely the outcome intervention is meant to avoid.

\subsection{Contact-rich Data Collection for Policy Training}
\label{sec:policy_data}
The user study measures what an operator can perceive and accomplish in a single session. We now ask whether that advantage persists once the human is out of the loop, by training policies on the data each interface produces. 
We evaluate on three contact-rich tasks (shown in Fig.~\ref{fig:tasks}) spanning distinct force-regulation demands. \textbf{Tong}: the robot picks up a pair of tongs, uses them to grasp an object, places the object at a target location, and returns the tongs to rest. The tool's aperture is controlled through sustained grip force, so the operator must modulate force finely enough to release the object without dropping the tool itself. \textbf{Raspberry}: the robot picks up a raspberry and places it in a glass bowl. The raspberry crushes under modest force, so the task tests grip regulation near the lower limit of a stable grasp, where vision gives no indication of how close the grasp is to failure. \textbf{Battery}: the robot picks up a battery and inserts it in a holder using a single hand. Insertion requires compressing a spring contact, so force must be applied along a specific direction while the grasp is maintained, exercising in-hand dexterity under load.

These tasks were also selected with the intervention setting in mind. In each, the robot holds an object or tool for most of the episode, so a takeover must preserve an existing grasp rather than begin from a free hand. A discontinuity at the moment of transfer leads to an unrecoverable state: dropped tongs, a dislodged battery, or a crushed raspberry. Tong can be generalized to other tool-use tasks, which we expect to be common when robots are deployed in human environments. A policy that begins to fail while holding a tool must be taken over smoothly, since losing the tool ends the episode in a state the policy cannot recover from. 

\textbf{Baselines, data collection, and policy training details.}
We compare against Manus Pro, serving as a pure-tracking baseline that isolates the contribution of the bilateral loop. Both interfaces are used identically throughout: for collecting the initial demonstrations, for supplying DAgger interventions, and for every condition we report. In this work, we use a Franka Panda robot for all data collection. To track the hand pose, we use a Meta Quest 3S headset, whose controller is mounted on the interface. We use two cameras for visual data recording, one Zed Mini on the wrist and one Zed 2 mounted on the left side of the table.

For each task we collect demonstrations through both interfaces under similar scene resets, identical task definitions, and demonstration counts: $80$ per interface for tong and $60$ per interface for raspberry and battery. 
A single operator collects all data for a given task across every condition, so differences between conditions are not confounded by operator skill.

All policies are trained with identical architecture and hyperparameters, and differ only in the demonstrations they consume. In this work, we use the UNet variant of diffusion policy~\cite{chi2023diffusion}. For observations, we include side-view and wrist-view images, the pose of the end-effector (e.e.), the joint angles of the robotic hand and tactile signal readings if provided. Tactile data are fed as low-dimensional observations, similar to joint angles and e.e. pose. We train all policies for 50 epochs, a number that we empirically found effective.
Detailed hyperparameters of our training will be publicized in our codebase.

Each policy is evaluated with $30$ rollouts from the same starting conditions, with object placements reset to a reference configuration verified against a fixed camera view, so comparisons across rounds and baselines are paired rather than independent. 
At least $10$ of the $30$ starting conditions for each task lie outside the distribution of object placements seen during pretraining, so the protocol measures generalization as well as in-distribution competence. Every rollout is scored against a fixed set of task stages, and a stage is credited only if that stage and all preceding stages succeeded, so the final stage is the task success rate. 

\subsection{DAgger with Human-Intervened Data}
\label{sec:dagger}

We run two rounds of DAgger per task and interface, collecting $20$ intervention trajectories per round. Each round begins with a full evaluation of the current policy, so the operator enters the collection phase having already observed how that policy fails rather than intervening blind. During collection, the policy is executed and the operator takes over at their own discretion, such as when a rollout enters a potentially unrecoverable state. Successful interventions are retained and the policies are finetuned on the accumulated set. Specifically, for all DAgger runs, we finetune the policy for $10$ epochs. Each interface supplies interventions only to its own policy, so each arm corrects its own failure distribution. For example, for the Tong task, we collect human-intervention data with the tong placed flat initially since the grasp in this configuration is usually unstable. 

The moment of transfer is where the two interfaces differ most. 
Under \OURS, the exoskeleton back-drives their fingers to the robot hand's current configuration before control is taken over, so the pose they assume at takeover is the pose the robot is already in.
Manus is passive and cannot place the operator's hand, so control transfers directly and the operator must reconcile their configuration with the robot's unaided. Every \OURS intervention reported below was collected under reverse teleoperation, which makes it a component of the DAgger results rather than only of the user study.

DAgger changes both how much data a policy sees and where it is collected. To separate the two, we train additional policies on the same number of demonstrations as DAgger round two ($120$ for tong and $100$ for raspberry and battery), collected non-iteratively from the original task distribution rather than from policy rollouts. Any remaining gap is attributable to the states the data covers rather than its volume.

\subsection{DITTO-WUJI and DITTO-INSPIRE}
The experiments so far use a single robot hand. Here, we show that \OURS can be used for different robotic hands. On Wuji we collect demonstrations for two multi-step tasks: \textbf{toy pick-and-place} and \textbf{drawer}. The drawer task exercises distinct parts of the hand within a single episode: the operator hooks the drawer open with the middle through pinky fingers, pinches the block between thumb and index, and closes the drawer with a fist. It tests whether the interface supports switching between grasp types mid-task, including one that relies on the coupled fingers alone. On Inspire we collect demonstrations for \textbf{bunny lifting}, picking up a soft toy.

\section{Results}

\subsection{User study results}

\begin{figure}[t]
    \centering
    \setlength{\abovecaptionskip}{2pt}
    \includegraphics[width=\linewidth]{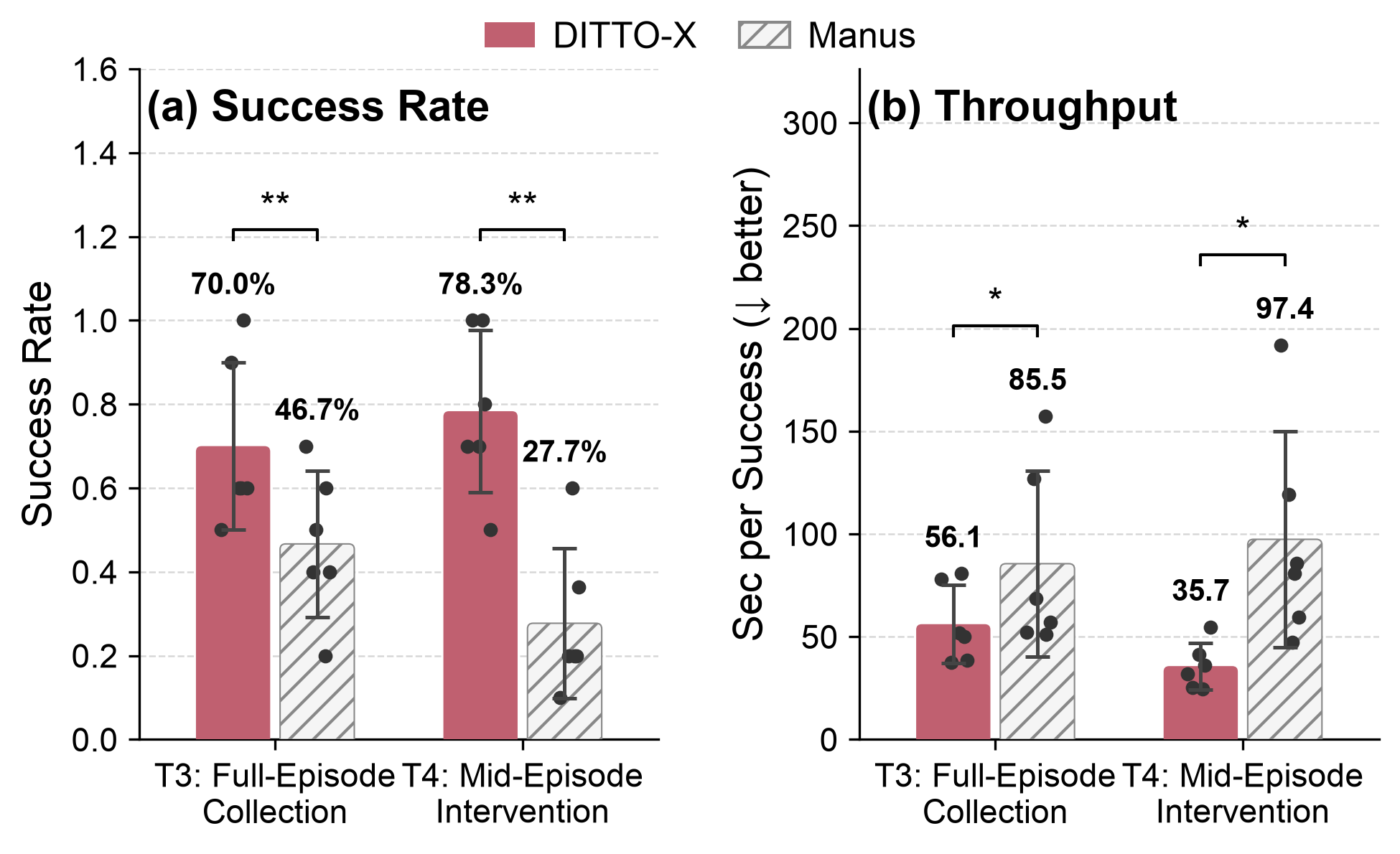}
    \caption{User study T3 and T4 success rates (a) and throughput (b). \vspace{-1em}}
    \label{fig:user_e4}
\end{figure}

\textbf{Object Size/Compliance Identification.}
With both feedback modes active, subjects discriminated object size correctly on $86.1\%$ of trials and compliance on $91.7\%$, well above the $33.3\%$ chance rate. Disabling either impairs both tasks: size falls to $61.1\%$ without vibrotactile feedback ($p=0.001$) and to $61.1\%$ without force feedback ($p=0.045$), while compliance falls to $80.6\%$ ($p=0.025$) and $47.2\%$ ($p=0.007$) respectively. Both are necessary, but not individually sufficient modes. Both properties are recovered without vision, which is the information an operator lacks when self-occlusion hides the hand during manipulation.

\begin{figure}[t]
    \centering
    \includegraphics[width= \linewidth]{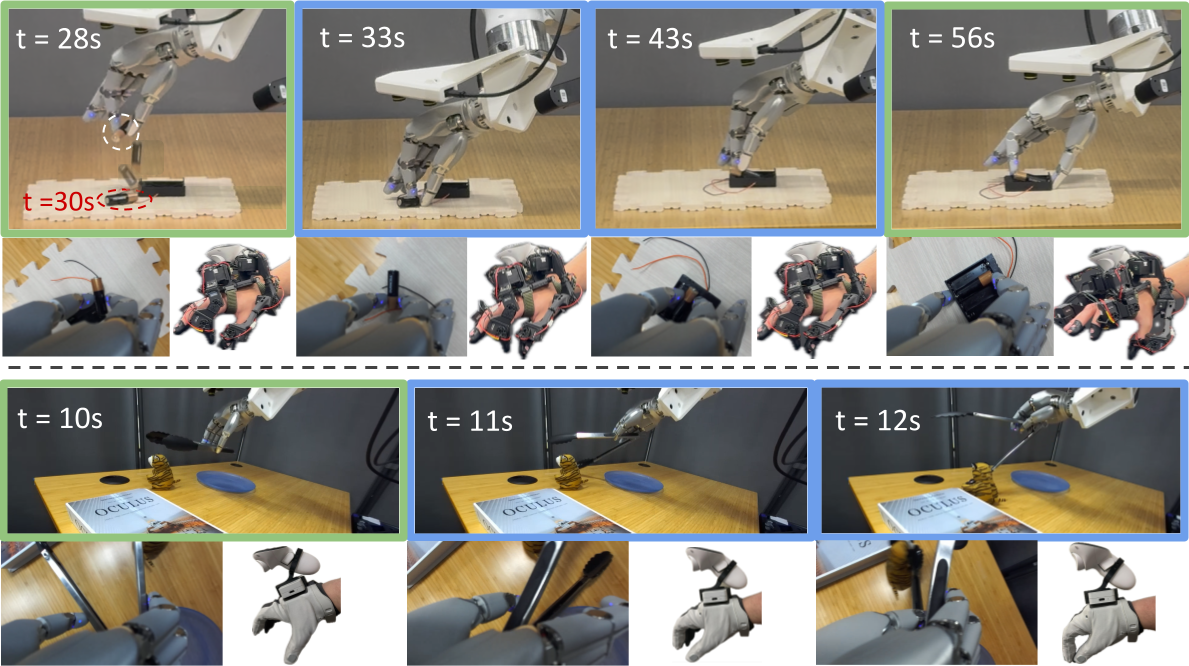}
    \caption{\textbf{Examples of human intervention with \OURS (Upper) and Manus (Bottom)}. The bottom row shows corresponding wrist camera views and operator hand pose. Human intervention is annotated with \textcolor{interventioncolor}{\textbf{blue}} border and policy execution is in \textbf{\textcolor{policycolor}{green}}.\vspace{-2em}}
    \label{fig:qualitative-human-intervention}
\end{figure}
\begin{figure}[t]
    \centering
    \setlength{\abovecaptionskip}{2pt}
    \includegraphics[width=0.95\linewidth]{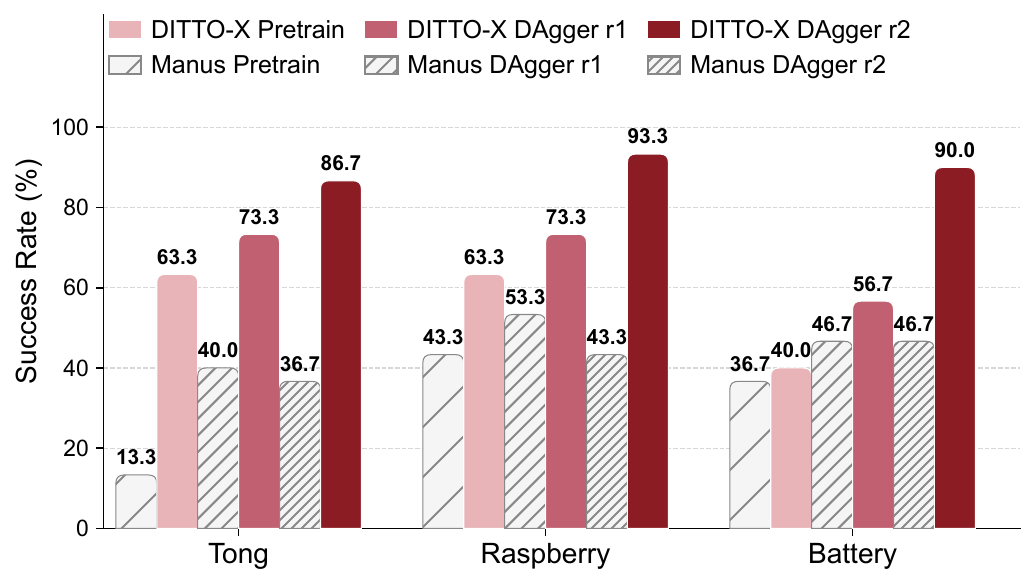}
    \caption{Policy success rates of pretrain and two rounds of DAgger.}
    \label{fig:dagger}
    \vspace{-2em}
\end{figure}

\textbf{Data collection and Policy Intervention (T3, T4).}
On data collection (T3), \OURS succeeded on $70.0\%$ of trials against $46.7\%$ for Manus, a within-subject gap of $23.3$ points that held for all six subjects (paired $t$-test, $t(5)=4.72$, $p=0.005$). Time per success fell from $85.5$\,s to $56.1$\,s, indicating that \OURS yields higher throughput for data collection. Notably, S3 had over 20 hours of prior experience controlling the Sharpa hand through Manus and no prior exposure to \OURS, yet still succeeded more often with \OURS.

On intervention (T4), our results show that subjects achieve much higher success rates (78.3\%$ \text{ v.s. } $27.7\%, $t(5)=4.45$, $p=0.007$), and time per success falls from $97.4$\,s to $35.7$\,s ($p=0.041$) using \OURS. This highlights that \OURS allows for robust robot-to-human authority transfer, which results in more completion of the full task.

Interestingly, our results show that subjects achieve higher success rates with the help of a policy (T4) than collecting full episodes (T3). In T4, the policy has already acquired the tool, so the operator completes a subset of T3, and a shortened task should be easier for both interfaces. For \OURS, success rate is higher when a subject is getting help from a policy. However, for Manus, due to failures in intervention, a policy's assistance does not improve overall task success.

Qualitatively (see Fig.~\ref{fig:candy}), in this task, the policy helps the subject with stable grasps, resulting in better release of the object without dropping the Tong. On the other hand, due to a worse transition from the robot to human, human-intervened execution does not outperform full trajectory collection. When using Manus (see Fig.~\ref{fig:qualitative-human-intervention}.Bottom), the operator tried to intervene on the robot as it was approaching the toy. Due to different hand pose between the human hand and the robotic hand, the grasp became unstable and the operator finally dropped the Tong.  Fig.~\ref{fig:qualitative-human-intervention} shows more qualitative examples of using \OURS for human intervention. For example, in the battery task, the battery slips out of the robotic hand during policy deployment ($t=28s$). Then the human operator assists the policy by picking up the battery and putting it in the battery holder ($33s\leq t \leq 43s$). Finally, the operator hand control back to the robot and the policy autonomously insert the battery into the holder.

\subsection{Policy Learning and DAgger Results}
\label{sec:policy_learning_results}

Figure~\ref{fig:dagger} reports success rates across DAgger rounds, and Table~\ref{tab:stages} breaks each rollout into task stages, where a stage counts only if every preceding stage succeeded. 

\textbf{Demonstration quality.}
Before any DAgger round, policies trained on \OURS demonstrations already outperform those trained through Manus on all three tasks. The margin is largest on tong, the task with the most demanding contact requirements: the operator must maintain a stable grasp on the tongs while using them to apply force to a second object, so any instability in the initial grasp propagates through the rest of the trajectory.

Quantitatively, Table~\ref{tab:stages} shows stage-wise success for each task. Both interfaces acquire the tongs at comparable rates, so neither is limited by reaching or closing on the tool. They diverge immediately afterward: \OURS picks up the toy on $66.7\%$ of rollouts against $23.3\%$ for Manus. What separates them is the stability of a grasp rather than its formation--the Manus policies close on the tongs without a secure hold, and the first contact with the toy displaces the tool before the toy is captured. The task fails where the tool must transmit force, not where it is picked up.

\textbf{Iterative correction.}
\OURS improves monotonically on every task (Fig.~\ref{fig:dagger}), while Manus improves at round one and then declines on two of three tasks, ending below its round-one performance on tong and raspberry. We attribute this to the transition itself: an intervention that begins with a mismatched hand pose produces a corrective trajectory that first perturbs the grasp and only then recovers it, and that trajectory becomes a training target. On tong, where the tool must be held stably throughout, such corrections teach the policy the very instability they were collected to fix.
\begin{figure}[t]
    \centering
    \includegraphics[width=0.95\linewidth]{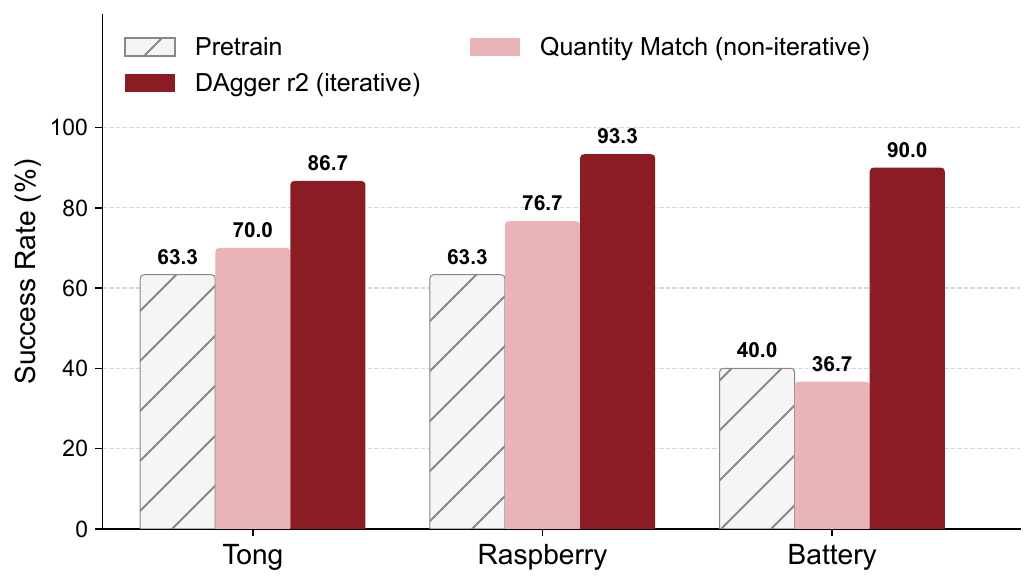}
    \setlength{\abovecaptionskip}{2pt}
    \caption{Comparisons between DAgger and quantity match baselines.}
    \label{fig:qty_baseline}
    \vspace{-5pt}
\end{figure}

\begin{table}[t]
\centering
\caption{Stage-wise completion rate (\%) over $30$ rollouts per run. A stage is credited only if all preceding stages succeeded.}
\label{tab:stages}
\footnotesize
\setlength{\tabcolsep}{4pt}
\begin{tabular}{llcccc}
\toprule
& & \multicolumn{2}{c}{\textbf{Pretrain}} & \multicolumn{2}{c}{\textbf{\OURS DAgger}} \\
\cmidrule(lr){3-4} \cmidrule(lr){5-6}
& & Manus & \OURS & r1 & r2 \\
\midrule
\multirow{4}{*}{\textbf{Tong}}
 & Grasp   & 73.3 & 76.7 & 90.0 & \textbf{96.7} \\
 & Toy     & 23.3 & 66.7 & 83.3 & \textbf{90.0} \\
 & Place   & 23.3 & 63.3 & 80.0 & \textbf{90.0} \\
 & Release & 13.3 & 63.3 & 73.3 & \textbf{86.7} \\
\cmidrule(lr){1-6}
\multirow{2}{*}{\textbf{Rasp.}}
 & Grasp   & 60.0 & 90.0 & \textbf{96.7} & 93.3 \\
 & Place   & 43.3 & 63.3 & 73.3 & \textbf{93.3} \\
\cmidrule(lr){1-6}
\multirow{3}{*}{\textbf{Battery}}
 & Grasp   & 70.0 & 80.0 & 93.3 & \textbf{100.0} \\
 & Align   & 46.7 & 56.7 & 80.0 & \textbf{93.3} \\
 & Insert  & 36.7 & 40.0 & 56.7 & \textbf{90.0} \\
\bottomrule
\vspace{-3.5em}
\end{tabular}
\end{table}

The stage breakdown also shows where \OURS gains come from. On battery, grasping and alignment are already largely reliable after round one and nearly all remaining failures occur at insertion; the second round is where insertion improves, from $56.7\%$ to $90.0\%$, closing most of the gap to the stage before it. Corrections gathered from a policy's failures repair those particular states rather than lifting performance evenly across the task.

\textbf{Data quantity vs DAgger.}
To separate iterative correction from data volume, we train a further policy on the same number of demonstrations as DAgger round two, collected non-iteratively (Fig.~\ref{fig:qty_baseline}). The matched baseline falls well short on all three tasks, and on battery it does not improve on the pretrained policy at all despite the additional demonstrations. Additional data drawn from the original distribution does not address the states where the policy fails; corrections collected at those states do. For example, on raspberry, we find that with DAgger, our policy are able to learn reach to out-of-distribution bowl locations, while the pretrain policy would drop the raspberry in wrong locations.

\textbf{Generalization across hands.}
Our results also show that \OURS can be used for collecting data for different hands.
Policies trained on the Wuji and Inspire demonstrations reach $83.3\%$ on toy pick-and-place, $66.7\%$ on drawer, and $93.3\%$ on bunny lifting, showing that \OURS collects usable data on hands of markedly different actuation. Reverse teleoperation is implemented for both, though we leave DAgger experiments on these hands to future work.

\textbf{Data from a proficient operator.}
For reference, we report the operator's own teleoperation success rate during data collection. This operator is proficient with both interfaces and distinct from the six novices in the user study. When collecting the pretraining demonstrations, $93.9\%$ of attempts through \OURS produced a usable trajectory against $71.7\%$ through Manus. When intervening on a running policy to collect DAgger data, the rates are $96.5\%$ and $58.2\%$. The performance difference increase at intervention for the expert as it does for novices in T3 and T4, suggesting the advantage does not diminish with familiarity.

\vspace{-3pt}
\section{Conclusions}
\label{sec:discussion}
We present \OURS, a dexterous teleoperation framework supporting both forceful teleoperation and fluent human intervention across diverse robotic hands, with multi-DoF force feedback and fingertip haptic feedback. Its actuated design enables \textit{reverse teleoperation}, letting the operator match the robot's state precisely and intervene smoothly in a learned policy. Our results show improved success rate and throughput for both regular teleoperation and human-in-the-loop policy deployment, and we believe the approach points toward a more efficient way to collect high-quality robot data.
\vspace{-9pt}
\section*{Acknowledgment}
We thank Han Zhang, Mengda Xu, Sirui Chen, and Sarthak Kamat for the insightful discussions and valuable feedback throughout this work. We also thank Sharpa for equipment support. 
\vspace{-3pt}

\bibliographystyle{IEEEtran}
\bibliography{bib/references}

\addtolength{\textheight}{-12cm}   
\end{document}